\documentclass{article}
\usepackage{spconf,amsmath,graphicx}
\usepackage{amssymb}
\usepackage{booktabs}
\usepackage{fixmath}
\usepackage{url}
\usepackage{float}
\usepackage[autostyle]{csquotes}

\title{A JOINT CONVOLUTIONAL AND SPATIAL QUAD-DIRECTIONAL LSTM NETWORK FOR PHASE UNWRAPPING}
\name{Malsha V. Perera $^*$, Ashwin De Silva $^*$ \thanks{$^*$ these authors contributed equally to the work. \newline © 2020 IEEE. Personal use of this material is permitted. Permission from IEEE must be obtained for all other uses, in any current or future media, including reprinting/republishing this material for advertising or promotional purposes, creating new collective works, for resale or redistribution to servers or lists, or reuse of any copyrighted component of this work in other works.}} 
\address{Dept. of Electronic and Telecommunication Eng., Univeristy of Moratuwa, Sri Lanka}

\begin{document}
%
\maketitle

\begin{abstract}
Phase unwrapping is a classical ill-posed problem which aims to recover the true phase from wrapped phase. In this paper, we introduce a novel Convolutional Neural Network (CNN) that incorporates a Spatial Quad-Directional Long Short Term Memory (SQD-LSTM) for phase unwrapping, by formulating it as a regression problem. Incorporating SQD-LSTM can circumvent the typical CNNs' inherent difficulty of learning global spatial dependencies which are vital when recovering the true phase. Furthermore, we employ a problem specific composite loss function to train this network. The proposed network is found to be performing better than the existing methods under severe noise conditions (Normalized Root Mean Square Error of $1.3 \%$ at $\text{SNR} = 0\text{ dB}$) while spending a significantly less computational time ($0.054 \text{s}$). The network also does not require a large scale dataset during training, thus making it ideal for applications with limited data that require fast and accurate phase unwrapping.
\end{abstract}

\section{Introduction}
\label{intro}
The problem of phase unwrapping is prevalent in many applications such as Quantitative Susceptibility Mapping (QSM) in Magnetic Resonance Imaging (MRI) \cite{MRI}, Synthetic Aperture Radar (SAR) interferometry \cite{SAR}, Fringe Projection Techniques (FPT) \cite{FTP} and digital holographic interferometry \cite{digiHolo}. Its objective is to recover the true phase signal from an observed wrapped phase signal that is in the range of $(-\pi, \pi]$. While it may be easy to recover true phase from wrapped phase under ideal conditions, phase unwrapping problem becomes challenging at the presence of noise, phase discontinuities and rapid variation of phase.

Phase unwrapping problem is traditionally addressed in two major approaches: path following approach and minimum norm approach. Path following approaches such as Quality-Guided Phase Unwrapping (QGPU) algorithm \cite{QGPU} and branch cut algorithm \cite{BranchCut} perform phase unwrapping by integrating phase along a selected path. Even though path following algorithms are relatively computationally efficient, they are not robust to noise. Minimum norm based algorithms \cite{min_norm1} are robust to noise, but they are computationally inefficient than path following approaches.

In recent years, deep learning algorithms have gained popularity and achieved state-of-the-art performance in many computer vision tasks. Following this trend, a few recent studies \cite{PhaseNet2,ZhangJ,ZhangT,Wang} have attempted to apply deep learning to address the phase unwrapping problem. Out of these, \cite{PhaseNet2,ZhangJ,ZhangT} have reformulated the phase unwrapping problem as a semantic segmentation task where Fully Convolutional Networks (FCNs) are trained to predict the wrap count at each pixel. Among these methods, Spoorthi et. al's \cite{PhaseNet2} PhaseNet 2.0, a deep encoder-decoder architecture comprised of dense blocks \cite{DenseNet}, has the best phase unwrapping performance. To the best of our knowledge only Wang et al \cite{Wang} has considered phase unwrapping as a regression problem where an FCN inspired by U-NET \cite{Unet} and ResNet \cite{ResidualNet} is employed to directly estimate the true phase from wrapped phase. Many of these FCN based phase unwrapping methods have achieved good performance under varying levels of noise while taking significantly less computational times, compared to traditional methods. Although there are benefits, these networks require large scale datasets, thus reducing their suitability in real-world applications.

In addition, the methods based only on FCNs pose another issue. The locally performed convolutional and pooling operations of typical CNNs often ignore the global spatial relationships between different regions of the image. Since most of the real-world phase images contain certain spatial structures, modelling such global spatial relationships is vital when learning the mappings from wrapped phase to true phase. Recurrent Neural Networks (RNNs) \cite{RNN} are a type of neural networks that can model the contextual relationships within a temporal sequence. However, it is not possible to directly apply an RNN to a feature map of an image. ReNet \cite{renet} and C-RNN \cite{C-RNN} introduce ways of applying RNNs to feature maps, and inspired by them, Ryu et al \cite{ryu} attempted to perform phase unwrapping in MRI images using a combination of a convolutional and recurrent networks. However, this work does not provide any quantitative result, and has not considered the influence of noise on phase unwrapping. Although conventional RNNs may result in some success, due to their limitation of modelling long-term dependencies of a sequence, they would be less suitable in modelling the spatial relationships within a lengthy sequence derived from a feature map. Long Short-Term Memory (LSTM)\cite{LSTM}, a special type of RNNs that is capable of modelling long term dependencies would be more adequate in this setting.

To address these shortcomings, in this paper, we propose an encoder-decoder CNN architecture that incorporates a Spatial-Quad Directional LSTM module which combines the power of FCNs and LSTMs in order to perform accurate and fast phase unwrapping, without having to train using a large scale dataset. We then describe a problem specific composite loss function comprised of variance of error and total variation of error losses to train this network. Finally, we report the findings of a comprehensive study which compared the proposed network with PhaseNet 2.0, Ryu et al's method and QGPU at varying levels of noise. These findings affirm that compared to other methods, the proposed network displays a strong robustness to severe noise conditions and a high computational efficiency when performing phase unwrapping.

\section{Methodology}
\label{method}

\subsection{Data Generation}
\label{data-generation-method}

Datasets used in this study are comprised of synthetic phase images containing random shapes, and their corresponding wrapped phase images. These random shapes are created by adding and subtracting several Gaussians having varying shapes and positions. Mixing Gaussians this way ensures that irregular and arbitrary shapes are formed instead of explicit patterns, which would, in return, enable the proposed network to learn the phase continuities pertaining to any general pattern. In addition, randomly selected slopes are also added to these synthetic phase images along vertical and horizontal directions in order to incorporate the ramp phase.

The wrapped phase image $\psi(x, y)$ of synthetic phase image $\phi(x, y)$ is computed as follows.
\begin{equation}
    \psi(x, y) = \angle \exp{\big(i \phi(x, y)\big)}
    \label{wrapping-equation}
\end{equation}
where, $(x, y)$ denotes the spatial coordinates of a pixel.

Following this method,  2 datasets,  each consisting of 6000 phase images ($256 \times 256$) with values ranging from  $-44$ to $44$ were created.  The phase images of only one of these datasets were randomly given additive Gaussian noise levels of $0, 5, 10, 20$ and  $60 \text{ dB}$ before wrapping them in order to simulate the noise prevalent in wrapped phase images of real-world applications. We shall denote this dataset as the \enquote{noisy} the other as  \enquote{noise free}.
 
\subsection{Spatial Quad-Directional LSTM Module}
\label{SQD-LSTM-block}

Let $\mathbf{X} = \{ x_{ij} \} \in \mathbb{R}^{w \times h \times c}$ be an input feature map, where $w$, $h$ and $c$ being the width, height and the feature dimensionality of $\mathbf{X}$. From $\mathbf{X}$, we can derive four distinct sequences as follows.
\vspace{-1mm}
\begin{align*}
    x_{\rightarrow} = \left\{\mathbf{r}_i\right\}_{i=1...h} ; \mathbf{r}_i=\left(x_{i1}, x_{i2},\ldots,x_{iw}\right)\\
    x_{\leftarrow} = \left\{\mathbf{r}_i\right\}_{i=h...1} ; \mathbf{r}_i=\left(x_{iw},\ldots,x_{i2},x_{i1}\right)\\
    x_{\downarrow} = \left\{\mathbf{r}_i\right\}_{i=1...w} ; \mathbf{r}_i=\left(x_{1i}, x_{2i},\ldots,x_{hi}\right)\\
    x_{\uparrow} = \left\{\mathbf{r}_i\right\}_{i=w...1} ; \mathbf{r}_i=\left(x_{hi},\ldots,x_{2i},x_{1i}\right)
\end{align*}

\begin{figure}[!t]
\centerline{\includegraphics[width=\columnwidth]{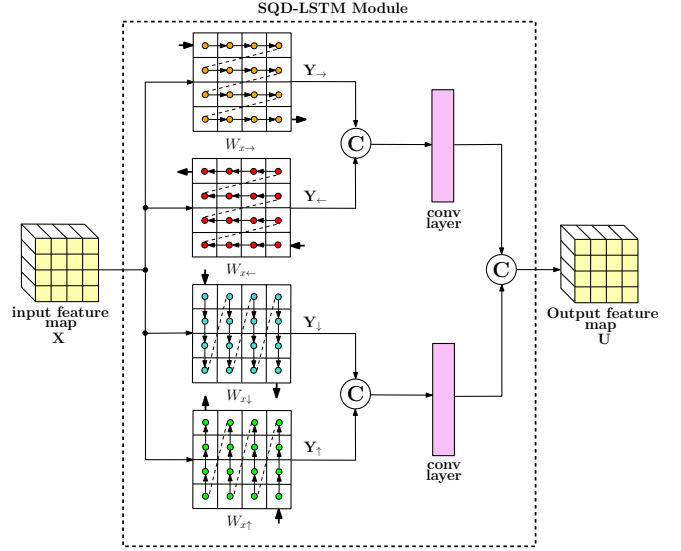}}
\caption{The proposed SQD-LSTM module}
\label{sqd-lstm}
\end{figure}

In other words, $x_{\rightarrow}$, $x_{\leftarrow}$, $x_{\downarrow}$ and $x_{\uparrow}$ represent the sequences obtained when one traverses $\mathbf{X}$ from left-to-right, right-to-left, top-to-bottom and bottom-to-top respectively as illustrated in Fig. \ref{sqd-lstm}. Let $x$ be any of the four sequences above. Then $x^{(s)} \in \mathbb{R}^c$ where $s \in \left[1, \dots, w \times h\right]$ is the feature vector that describes a corresponding region of the original image.

In order to model the spatial relationships within these four sequences, we utilize LSTMs which are specialized in modelling long-term dependencies. In simple terms, the input-output relationship of an LSTM layer for sequence $x$ can be written as follows. 

\begin{equation}
    y^{(s)} = \mathbb{LSTM}\big( x^{(s)}, y^{(s-1)} ; W_x, u\big)
\end{equation}

where, $y^{(s)} \in \mathbb{R}^u$ is the LSTM output at $s$, $W_x$ is the unified LSTM weights and $u$ is the number of units in the layer. In our SQD-LSTM module, the $x_{\rightarrow}$, $x_{\leftarrow}$, $x_{\downarrow}$ and $x_{\uparrow}$ are processed by 4 separate LSTMs with unified weights  $W_{x_{\rightarrow}}$, $W_{x_{\leftarrow}}$, $W_{x_{\downarrow}}$ and $W_{x_{\uparrow}}$, respectively. By rearranging the output sequences $y_{\rightarrow}$, $y_{\leftarrow}$, $y_{\downarrow}$ and $y_{\uparrow}$ from the 4 LSTMs, we obtain the output feature maps $\mathbf{Y}_{\rightarrow}$, $\mathbf{Y}_{\leftarrow}$, $\mathbf{Y}_{\downarrow}$ and $\mathbf{Y}_{\uparrow}$ each in $\mathbb{R}^{w \times h \times u}$. 

The output feature maps pertaining to horizontal ($\mathbf{Y}_{\rightarrow}$, $\mathbf{Y}_{\leftarrow}$) and vertical ($\mathbf{Y}_{\downarrow}$ , $\mathbf{Y}_{\uparrow}$) directions are concatenated as illustrated in Fig. \ref{sqd-lstm} and are passed through separate $3 \times 3$ convolutional layers containing $d$ filters each, in order to explore the local association of the spatial features obtained from the LSTM layers. Finally, the output of the SQD-LSTM $\mathbf{U} \in \mathbb{R}^{w \times h \times 2d}$ is formed by concatenating the output feature maps from the two convolutional layers. In this study, $u$ and $d$ were set to $32$ and $64$ respectively.

\vspace{-3mm}
\subsection{Network Architecture}
\label{data-generation-method}

\begin{figure}[!t]
\centerline{\includegraphics[width=\columnwidth]{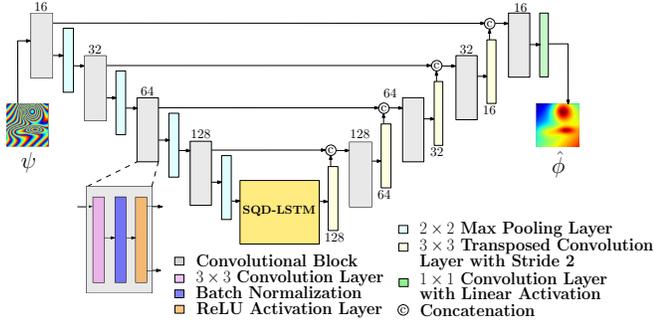}}
\caption{The proposed network architecture. The number of filters in each convolutional block is stated above it and the number of filters in each transposed convolution layer is stated below it.}
\label{Network}
\end{figure}

The proposed network architecture is illustrated in Fig. \ref{Network}. In a nutshell, the architecture is comprised of a fully convolutional encoder-decoder network where the output of the encoder is passed through the proposed SQD-LSTM module before fed in to the decoder. The output feature map of the encoder is able to represent the local information of the input image. Feeding this encoder output to the SQD-LSTM module allows the network to learn the spatial dependencies between the local features contained in the encoder output. Subsequently, the output of SQD-LSTM module is fed in to the decoder network which increases the resolution of the output through transpose convolutional operations. Furthermore, in order to combine semantic features from the decoder layers and local features from the encoder layers, we add skip connections as shown in Fig. \ref{Network}. Adding skip connections this way, ensures that the network assembles a more refined output at the later layers.

Every convolutional block in the network contains a $3 \times 3$ convolutional layer which is followed by a batch-normalization layer and a ReLU activation. Every encoder convolutional block is followed by a $2 \times 2$ max-pooling layer with a stride of $2$, while every decoder convolutional block is preceded by $3 \times 3$ transposed convolutional layers with a stride of $2$. Since the network performs a regression task, the final convolutional block of the decoder layer is succeeded by a $1 \times 1$ convolutional layer with a linear activation.

\subsection{Loss Functions}
\label{loss}

Since we formulate the phase unwrapping problem as a regression task, the go to choice for the loss function is Mean Squared Error (MSE) loss. However, our experiments reveal that MSE loss, when employed to the proposed network, shows an insufficient convergence that results in poor performance of phase unwrapping. By \eqref{wrapping-equation}, it follows that $\phi + 2\pi n$ where $\forall n \in \mathbb{Z}$ gives rise to the same wrapped phase $\psi$. Therefore, the phase unwrapping problem of $\psi$ does not have a unique solution. Since MSE loss enforces the network to learn a unique solution, it does not fit well in to the nature of phase unwrapping problem. Hence, a loss function which allows for other solutions at convergence while increasing the similarity between the predicted phase $\hat{\phi}$ and true phase $\phi$ is required. To address these concerns, we adopt the composite loss function $\mathcal{L}_c$ defined below.
\begin{equation}
    \mathcal{L}_c = \lambda_1 \mathcal{L}_{var} + \lambda_2 \mathcal{L}_{tv} 
\end{equation}
where,
\begin{equation}
    \mathcal{L}_{var} = \mathbb{E}\big[(\hat{\phi} - \phi)^2\big] - \big(\mathbb{E} \big[(\hat{\phi} - \phi) \big] \big)^2
\end{equation}
\begin{equation}
    \mathcal{L}_{tv} = \mathbb{E}\big[|\hat{\phi}_x - \phi_x| + |\hat{\phi}_y - \phi_y|\big]
\end{equation}
and $\lambda_1$, $\lambda_2$ are the weights assigned for the two losses and were empirically set to 1 and 0.1 respectively during training. The variance of error loss $\mathcal{L}_{var}$ allows for alternate solutions at convergence while the total variation of error loss $\mathcal{L}_{tv}$ increases the increasing the similarity between $\hat{\phi}$ and $\phi$ by enforcing the network to match the gradients of them.

\section{Experiments and Results}
\label{experiments and results}

\begin{figure*}[htbp]
\centerline{\includegraphics[width=\textwidth]{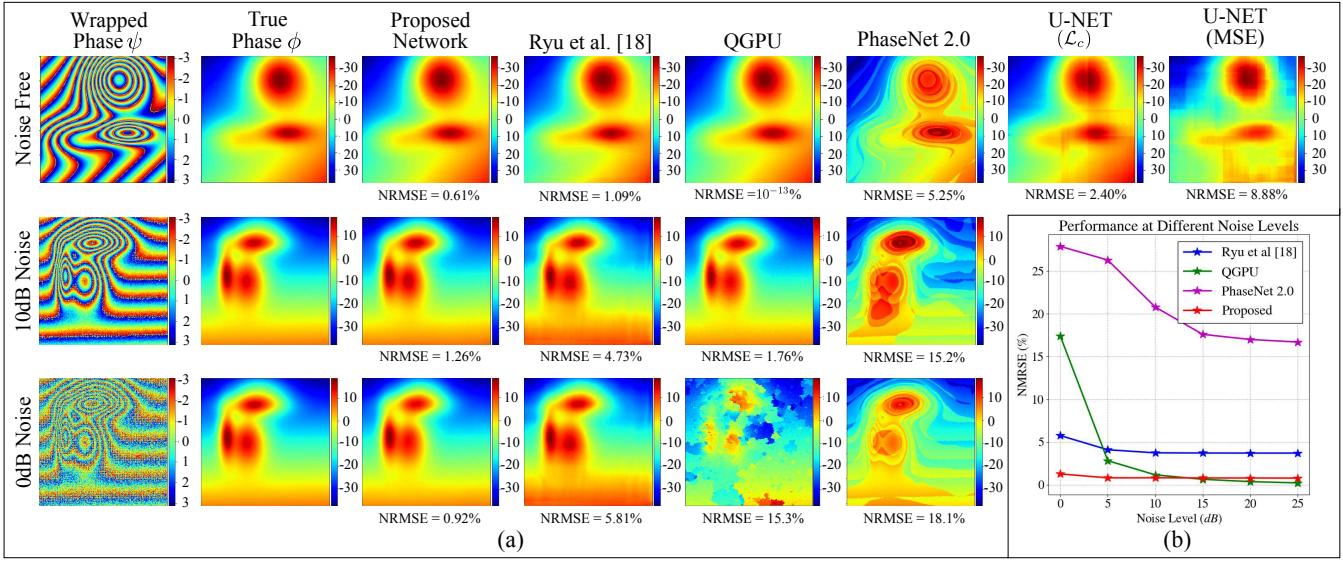}}
\caption{ (a) Unwrapped phase images obtained from different methods for selected noise free and noisy (10 $\text{dB}$ and 0 $\text{dB}$) wrapped phase images. (b) Plot of NRMSE of proposed method, Ryu et al \cite{ryu}, QGPU and PhaseNet 2.0  against varying noise levels.}
\label{results}
\end{figure*}

The proposed network was implemented in Keras  and was trained and tested on the two datasets mentioned in section \ref{data-generation-method}, separately. In both occasions, the models were trained using ADAM optimizer with a learning rate of 0.001 and they converged within 10 epochs taking only $\sim 1.5$ hours to train. Similarly,  Ryu et al's \cite{ryu} network, PhaseNet 2.0 \cite{PhaseNet2} and QGPU \cite{QGPU} were implemented and tested on the two datasets as well. Out of these, Ryu et al's network and PhaseNet 2.0 were trained on both noisy and noise free datasets. Furthermore, in order to assess the significance of SQD-LSTM module and the loss function $\mathcal{L}_c$, we trained and tested two separate U-NETs (which resemble the convolutional  architecture of the proposed network), one with MSE and one with $\mathcal{L}_{c} $ as the loss function, using only the noise free dataset. All the above training and testing were performed on an NVIDIA Tesla K80 GPU with a fixed train - test split of 5000 - 1000 for each dataset.

To evaluate and compare these methods, we computed Normalized Root Mean Square Error (NRMSE - normalized by the range of the respective true phase image) of the unwrapped phase images and measured the average computational time per output, for each method. These results are summarised in Table  \ref{performance}.  Several selected unwrapped phase images obtained from each method are illustrated in Fig. \ref{results} together with a plot of the noise levels versus NMRSE.

\begin{table}[!h]
\vspace{-4mm}
\setlength\tabcolsep{0pt}
\caption{Results} \label{results table}
\centering
\smallskip
\begin{tabular*}{\columnwidth}{@{\extracolsep{\fill}}l cccc}
\toprule
Method & {Noise Free}& {Noisy} & {Computational} \\
 & {NRMSE}& {NRMSE} & {Time (s)} \\
 \midrule
  UNET (MSE)& $14.24\%$   & -     & $0.234$   \\
  UNET ($\mathcal{L}_{c} $)& $2.75\%$ & - & $0.262$    \\
\midrule
  Ryu et al.\cite{ryu} & $2.23\%$  & $3.84\%$  & $0.687$     \\
  PhaseNet 2.0 \cite{PhaseNet2} &  $9.41\%$  & $17.53\%$     & $0.234$    \\
  QGPU \cite{QGPU} & $\mathbf{10^{-13}\textbf{\%}}$ & $5.04\%$ & $35.42$  \\
  Proposed Method & $0.84\%$  & $\textbf{0.90\%}$ & $\textbf{0.054}$     \\
\bottomrule
\label{performance}
\end{tabular*}
\vspace{-8mm}
\end{table}
It follows from Table \ref{performance} that QGPU has achieved a nearly perfect performance in unwrapping noise free images. However, it exhibits poor performance when it comes to noisy images and has a very high average computational time ($35.42$ s) compared to deep learning approaches. The proposed method, on the other hand achieves a comparable performance (NRMSE = $0.84 \%$)  with that of QGPU in terms of noise free images, while attaining the highest performance (NRMSE = $0.9 \%$) in terms of noisy images and the lowest average computational time ($0.054$ s) among the compared methods. As evident from Fig. \ref{results} (b), the proposed network is able to accurately (NRMSE = $1.26 \%$)  unwrap the wrapped phase images with severe noise levels as high as $\text{SNR} = 0 \text{ dB}$. It is also observed that the proposed network has surpassed the PhaseNet 2.0 which currently holds the state-of-the-art deep learning based phase unwrapping performance. PhaseNet 2.0, is a comparatively deeper network comprised of dense blocks and therefore, its training process is highly data intensive. Nevertheless, in this study, its phase unwrapping error has increased, and this may be due to the relatively smaller size ($5000$) of the training dataset used. While being trained only on 5000 images, the proposed network however attained the highest phase unwrapping performance for noisy data among the considered methods, thus making it ideal for real-world applications with limited availability of data.

As shown by Table \ref{performance}, the U-NET with $\mathcal{L}_{c} $ has a better phase unwrapping performance compared to the U-NET with MSE. Therefore, it is apparent that $\mathcal{L}_{c} $ is a better suited loss function for this problem than MSE. Also, it is evident that the proposed method performs better than the U-NET with $\mathcal{L}_{c} $. During experimentation, we also noticed that the proposed network converges faster than any of the considered methods. These observations point to the conclusion that the success of the proposed network is owed to the the SQD-LSTM module and the $\mathcal{L}_{c} $ loss function.

\section{Conclusion}
\label{conclusion}

In this paper, we propose a novel convolutional architecture that encompasses a spatial quad-directional LSTM for phase unwrapping, by formulating it as a regression problem. When compared with several existing phase unwrapping methods, it is found that the proposed network attains state-of-the-art unwrapping performance even under sever noise conditions without requiring large scale datasets to train. Furthermore, the network expends a significantly less computational time on average to perform this task, making it ideal for applications which demand accurate and faster phase unwrapping. Looking forward, we hope to extend the network to address application specific challenges such as phase discontinuities.




\bibliographystyle{IEEEbib}
\bibliography{refs}

\end{document}